%% file: acl_latex.tex
\pdfoutput=1

\documentclass[11pt]{article}

\usepackage[final]{acl}
\usepackage[normalem]{ulem} 
\usepackage{times}
\usepackage{latexsym}

\usepackage[T1]{fontenc}
\usepackage{soul}
\usepackage[utf8]{inputenc}

\usepackage{microtype}

\usepackage{inconsolata}

\usepackage{graphicx}

\usepackage{enumitem}

\usepackage{booktabs}
\usepackage{multirow}

\usepackage{bbm}
\usepackage{amsmath}

\newcommand{\SYSNAME}{\textit{CrEst}}

\title{CrEst: Credibility Estimation for Contexts in LLMs via Weak Supervision}

\author{Dyah Adila\thanks{Work done during internship at AWS} \\ University of Wisconsin-Madison \\ \texttt{adila@wisc.edu}  \And Shuai Zhang  \\
  AWS AI Labs \\
  \texttt{shuaizs@amazon.com}
  \AND 
  Boran Han \\ AWS AI Labs \And Bonan Min \\ AWS AI Labs \And Yuyang Wang \\ AWS AI Labs}

\begin{document}
\maketitle

\input{latex/sections/abstract}

\input{latex/sections/introduction}

\input{latex/sections/method}

\input{latex/sections/experiments}

\input{latex/sections/analysis}

\input{latex/sections/related_work}

\input{latex/sections/conclusion}
\section*{Limitations}

$\SYSNAME$ relies on the assumption that credible documents agree with others in the retrieved set. This can fail if most retrieved documents are unreliable—e.g., due to adversarial noise or domain bias. While rare in practice, such cases may lead to inflated credibility scores for untrustworthy sources.

Our method also depends on the choice and quality of embedding models. Though ensembling improves robustness, the gains plateau beyond a few embedders, and performance may vary across domains.

Finally, while the white-box variant allows fine-grained attention control, it requires access to model internals. The black-box variant is broadly applicable but incurs higher inference cost due to repeated prompting.

\section*{Ethics Statement}
We have carefully considered the ethical implications of our work and do not foresee any significant risks. $\SYSNAME$ is designed as a lightweight, annotation-free framework that does not require personal or sensitive data, which helps reduce concerns around privacy and data misuse.

Our method operates entirely on retrieved text from existing datasets and does not introduce new data collection pipelines. Furthermore, our reliance on publicly available datasets and open-source models promotes transparency, reproducibility, and responsible research practices. As $\SYSNAME$ aims to improve robustness to unreliable information, we believe it may contribute positively to mitigating misinformation in retrieval-augmented systems.



\bibliography{custom}

\appendix



\input{latex/appendix/glossary}

\input{latex/appendix/eq3_derivation}

\input{latex/appendix/dataset_details}

\input{latex/appendix/embedder_details}

\end{document}

%% file: latex/sections/abstract.tex
\begin{abstract}
The integration of contextual information has significantly enhanced the performance of large language models (LLMs) on knowledge-intensive tasks. However, existing methods often overlook a critical challenge: the credibility of context documents can vary widely, potentially leading to the propagation of unreliable information. In this paper, we introduce $\SYSNAME$, a novel weakly supervised framework for assessing the credibility of context documents during LLM inference—without requiring manual annotations. Our approach is grounded in the insight that credible documents tend to exhibit higher semantic coherence with other credible documents, enabling automated credibility estimation through inter-document agreement. To incorporate credibility into LLM inference, we propose two integration strategies: a black-box approach for models without access to internal weights or activations, and a white-box method that directly modifies attention mechanisms. Extensive experiments across three model architectures and five datasets demonstrate that $\SYSNAME$ consistently outperforms strong baselines, achieving up to a 26.86\% improvement in accuracy and a 3.49\% increase in F1 score. Further analysis shows that $\SYSNAME$ maintains robust performance even under high-noise conditions.
\end{abstract}

%% file: latex/sections/introduction.tex
\section{Introduction}

Large language models (LLMs) have achieved impressive results on knowledge-intensive tasks by utilizing external context during inference, setting a new benchmark for performance \cite{guu2020retrieval, lewis2020retrieval}. In this setup, context documents from a knowledge source are provided as in-context examples to guide the model's response. However, this approach hinges on a critical yet often overlooked assumption: the trustworthiness of the supplied content. When these documents contain irrelevant, misleading, or non-credible information, the LLM may generate responses that reflect or even amplify such inaccuracies \cite{xiang2024certifiably, zou2024poisonedrag}. In this paper, we address this challenge by introducing a mechanism that enables LLMs to assess and incorporate the credibility of each context passage during generation.

\input{./latex/figures/main}

Existing research has shown that incorporating manual credibility annotations can significantly improve performance \cite{pan2024contexts}. However, such annotations are resource-intensive, difficult to scale, and fail to generalize to large datasets. \citet{yan2024corrective} proposed combining documents from unrelated queries to differentiate relevant from irrelevant signals, but their method does not directly assess credibility. Other approaches use open-source re-rankers or instruction-tuned LLMs to generate relevance assessments \cite{deng2024cram, wei2024instructrag, wang2024astute}, but these rely heavily on the LLM’s intrinsic reasoning capabilities—an inherent limitation when deploying smaller or less powerful models. Meanwhile, \citet{xiang2024certifiably} recommend aggregating answers by querying each context separately, though this significantly increases computational overhead and API costs.

To overcome these limitations, we introduce $\SYSNAME$, a weakly-supervised framework for estimating document credibility without manual annotations. Our approach builds on the key insight that credible documents typically exhibit stronger semantic similarities with other documents within the in-context documents set. This principle draws inspiration from weak supervision  \cite{ratner2017snorkel, ratner2016data},  traditionally used for automatically labeling training data by aggregating noisy \emph{labeling functions} \cite{fu2020fast, shin2021universalizing}. We reformulate the credibility estimation problem within this framework by treating each context document as a labeling function and measuring their mutual agreements through semantic similarity in the latent space. To integrate these credibility estimates into LLM inference, we develop two complementary mechanisms: a black-box approach for models with restricted access to their weights and activation values, and a white-box method for models allowing direct manipulation of internal activations.

Our extensive experimental results demonstrate that $\SYSNAME$, despite being fully unsupervised and independent of model reasoning capacity, significantly outperforms strong baselines by up to 26.86\% in accuracy and 3.49\% in F1 score across three model architectures and five datasets. Notably, $\SYSNAME$ maintains its performance advantages even under stress tests with artificially increased noise in the contextual documents. 

\paragraph{Summary of contributions.}
\begin{itemize}[noitemsep, topsep=0pt, leftmargin=*]
\item $\SYSNAME$, a novel weak-supervision framework that estimates retrieved context credibility without manual annotations or model-generated signals.
\item We propose two integration strategies for incorporating credibility estimates into LLM inference: \ul{\textit{one tailored for black-box models and another for white-box models}}.

\item Comprehensive empirical validation across multiple architectures and datasets, demonstrating robust performance improvements even under high-noise conditions
\end{itemize}

%% file: latex/figures/main.tex
\begin{figure*}[htp!]
    \centering
    \includegraphics[width=\linewidth]{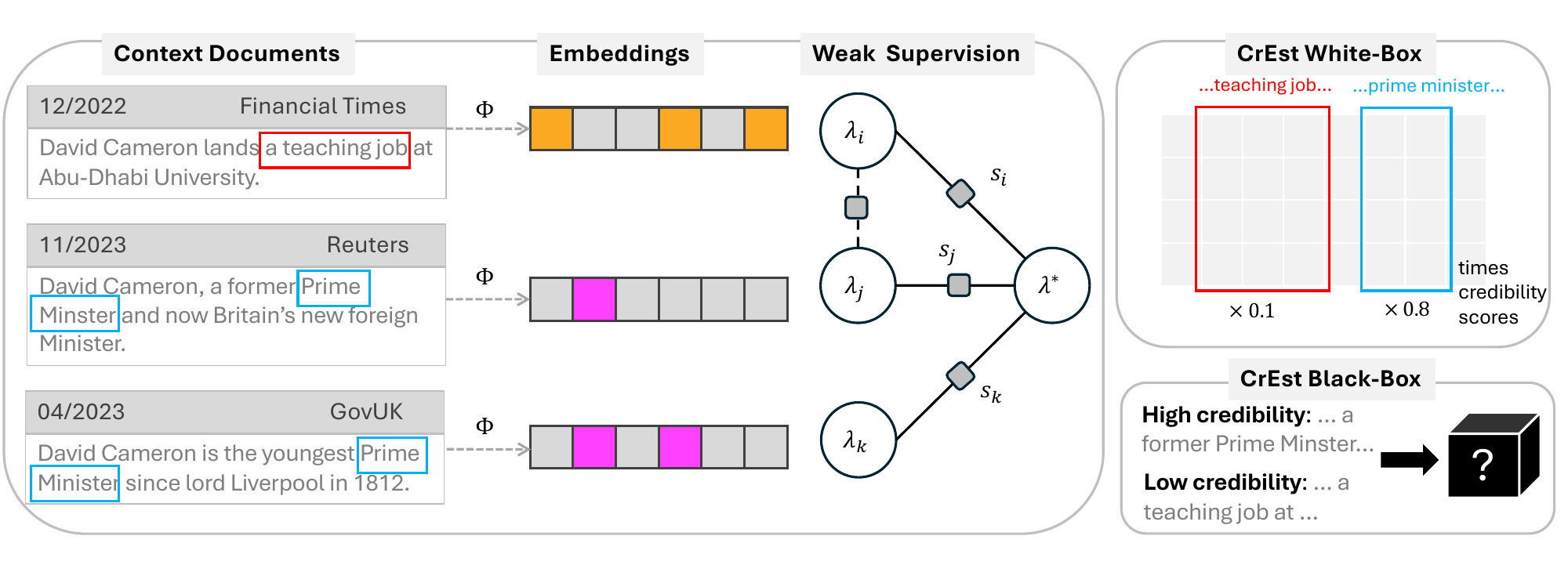}
    \caption{First, documents are embedded into the latent space. Next, we measure the pairwise distance between the context documents; using weak supervision to estimate each document's credibility, given the pairwise distances. The scores are integrated using two mechanisms: one tailored for white-box models with transparent internals and another for black-box models limited to input-output access.}
    \label{fig:main}
\end{figure*}

%% file: latex/sections/method.tex
\section{The Proposed Method: $\SYSNAME$}

We now formally describe the $\SYSNAME$ algorithm. Given a user query $x$, a retriever is used to obtain a set of relevant documents $\mathcal{D} = \{d_1, \ldots, d_n\}$. Next, an LLM $\mathcal{M}$ is tasked to generate a prediction $\hat{y}$ when prompted with $x$ concatenated with $\mathcal{D}$ as in-context samples, formally written as $\hat{y} = \mathcal{M}(x, \mathcal{D})$. 

Our approach, $\SYSNAME$,  first estimates the credibility scores of the retrieved documents $\mathcal{S} = \{s_1, \ldots, s_n\}$.  We then combine the documents $\mathcal{D}$ and their scores $\mathcal{S}$ with the input $x$ to create an augmented input, formally written as $\mathcal{M}(x, \mathcal{D}, \mathcal{S})$. The augmented input can be in the form of a prompt with the scores incorporated as prompting text (i.e., \textit{{\text{low}, \text{medium}, \text{high}} credibility}), or by directly using the scores to edit the attention mask during inference. 

In the following, we first provide the necessary background on weak supervision in Section~\ref{sec:ws_background}, followed by a detailed description of the credibility estimation algorithm in Section~\ref{sec:method_cred_estimation}. We then explain how our approach integrates these credibility scores into the LLM inference pipeline in Section~\ref{sec:black_white_box}. The end-to-end pipeline of $\SYSNAME$ is illustrated in Figure~\ref{fig:main}.

\subsection{Preliminary: Weak Supervision}
\label{sec:ws_background}

We model the relationship between each query's retrieved documents $\{d_1, \ldots, d_n\}$ with an unobserved `\textsc{true}' document $d^*$ as probabilistic graphical model. We assume that there exists an unobserved `\textsc{true}' document whose difference with all other retrieved documents is small; in other words it is close to most of the retrieved documents and contradicts very few, if any, of them. Since measuring the difference at the token level is intractable, we instead assess semantic differences between documents in the embedding space.

Let $\Phi$ be an embedding function that maps a document to a high dimensional space $\Phi : d_i \rightarrow z_i\in \mathbbm{R}^m$, where $m$ is the embedding dimension. We denote document $d_i$ embedding as $\lambda^i := \Phi(d_i)$ and the unobserved true document embedding as $\lambda^*$. We adopt the approach in \cite{shin2021universalizing} to model the distribution over embedding vectors as
\begin{equation}
\label{eq:graphical_model}
    \text{Pr}(\lambda^1,\ldots,\lambda_n | \lambda^*) = \frac{1}{Z}\text{exp}(-\sum_{i=0}^{n}\theta_i \| 
\lambda^i - \lambda^* \|^2),
\end{equation}
where $Z$ is the log partition functions, and $\theta = [\theta_1,\ldots,\theta_n]^T$ are canonical parameters of the graphical model. 

This model works as follow: if $\theta_i$ is large, any disagreement between $\lambda^*$ and $\lambda^i$ is penalized, resulting in a low probability. Thus $\lambda^i$ is unlikely to be very different from $\lambda^*$. Conversely, if $\theta_i$ is small,  such disagreements will be common; and $\lambda^i$ is likely to be very different from $\lambda^*$.

\subsection{Credibility Estimation with Weak Supervision}
\label{sec:method_cred_estimation}
Now, we introduce how to estimate documents credibility scores $\mathcal{S} = \{ s_1, \ldots, s_n \}$ using the graphical model in Equation \ref{eq:graphical_model}. Recall that we have access to the embeddings $\lambda^i$ of the retrieved documents, and that the unobserved true document embedding $\lambda^*$ has high similarity to most of the retrieved documents. Consequently, a document that is close in distance to $\lambda^*$ is also likely to be close to many other retrieved documents. Based on this idea, we define each document's credibility score as follows:
\begin{equation}
\label{eq:s_defiition}
s_i := \frac{1}{\mathbbm{E}[\| 
\lambda^i - \lambda^* \|^2]}
\end{equation}
This formulation indicates that the closer a document's embedding is to the true document's embedding—and thus to the embeddings of other retrieved documents—the higher its credibility score is. This leads us to our central assumption: \textit{a more credible document is one that aligns more closely with the majority of other retrieved documents}.

However, this assumption also highlights a limitation of our method: when more than half of the retrieved documents are unreliable, our approach may incorrectly assign high credibility scores to these unreliable but semantically similar documents. Fortunately, recent large-scale analyses of retrieval systems suggest that such cases are rare in practice, with the majority of retrieved documents maintaining reasonable reliability \cite{lioma2016study}. We elaborate on this limitation in Section \ref{sec:exp_noise}.

\paragraph{Computing $s_i$} From the graphical model in Equation \ref{eq:graphical_model}, it follows that for any $i,j \in [n]$
\begin{equation}
\label{eq:exppectation}
    \mathbbm{E}[\| \lambda^i - \lambda^j \|^2] = \mathbbm{E}[\| \lambda^i - \lambda^* \|^2] + \mathbbm{E}[\| \lambda^j - \lambda^* \|^2].
\end{equation}


Applying this equation to pairs of document embeddings over a triplet $\lambda^i, \lambda^j, \lambda^k$ yields a system of three equations with three unknown expectation terms~\citep{fu2020fast}. By solving this system, we obtain:
\begin{equation}
\label{eq:triplet}
    \mathbbm{E}[\| \lambda^i - \lambda^* \|^2] = \frac{1}{2}\delta_{ij}+\delta_{ik}-\delta_{jk} , \forall (i,j,k) \in [n], 
\end{equation}
where $\delta_{ij} =  \mathbbm{E}[\| \lambda^i - \lambda^j \|^2]$. We provide detailed derivation in Appendix \ref{sec:triplet_derivation}.

\paragraph{Denoising $s_i$ Estimates} The expectation terms in $\delta_{ij}$, $\delta_{ik}$, and $\delta_{jk}$ are influenced by the inherent randomness in their respective $\lambda^i$, $\lambda^j$, $\lambda^k$ variables. This stochastic nature can introduce estimation noise, particularly when working with a small set of retrieved documents, due to imperfections in the embedding function. To improve robustness, we employ an ensemble approach: computing credibility scores using multiple distinct embedding functions and aggregating their results to reduce variance in the final estimates.

Given $M$ embedding functions, each embedder calculates a document credibility score, resulting in $M$ separate sets of credibility scores. We denote these M sets of credibility scores as $\{\mathcal{S}^1, \ldots, \mathcal{S}^M\}$, where each $\mathcal{S}^m = \{s_1^m,\ldots,s_n^m\}$ represents the scores computed using the $m$-th embedding function. In the next section, we will describe how we aggregate these scores.

\subsection{Applying Credibility Estimates in Black- and White-Box Settings}
\label{sec:black_white_box}
Once we have document credibility estimates, we incorporate them into the LLM inference pipeline. The most straightforward method is to prompt LM with each document alongside its estimated credibility, expressed as $\hat{y} = \mathcal{M}(x, \mathcal{D}, \mathcal{S})$. However, this limits us to using only a single set of credibility scores, which may be noisy. 

We present two approaches for integrating credibility scores: the first is designed for black-box models, where model internal activations are inaccessible during inference; the second targets white-box models, allowing direct manipulation of activation values using credibility scores.

\paragraph{Multiple Prompting (Black-Box Model)} In black box model we only have access to modify its input (the prompt). To use multiple set of credibility scores, we prompt the model multiple times in parallel, each time with a different set of scores $\mathcal{S}^m, m \in [M]$, resulting in $M$ different outputs from the LM $\hat{y}_m = \mathcal{M}(x, \mathcal{D}, \mathcal{S}^m)$. We then aggregate these outputs by selecting the most popular one -- essentially, the output that aligns best with the majority of the other outputs. Notice that this notion is similar our intuition in deriving document credibility scores in Section \ref{sec:method_cred_estimation}. 

To determine output popularity, we use the same method as for calculating credibility scores. Specifically, given $M$ outputs $\{\hat{y}_1, \ldots, \hat{y}_M\}$, we first obtain their embeddings $\{\lambda^1, \ldots, \lambda^M\}$. We then calculate their popularity scores $\{p_1, \ldots, p_M\}$ using Equation \ref{eq:s_defiition} and select the output with the highest score, formally expressed as:
\begin{align*}
    \hat{y}^* = \hat{y}_{\mathop{\arg\max}_m}(p_m).
\end{align*}

\paragraph{Attention Mask Editing (White-Box Model)} Alternatively, we can use $\mathcal{S}$ to directly modify the LM's attention mask values. This allows us to guide the model on which documents it should prioritize or downplay in its processing. Previous studies \citep{zhang2023tell, deng2024cram,adila2024discovering} have demonstrated that modifying the attention scores of specific input tokens can effectively change the model's output, making it more or less influenced by those tokens. 

To calculate the final score used to scale the attention mask, we simply aggregate the estimated credibility scores from multiple $M$ embedders $\mathcal{S}^1,\ldots,\mathcal{S}^M$ by taking their average. To avoid one set of score dominating due to larger score scale, we standardize them to 0-1 scale before taking their average. Each document $d_i$ aggregated score is computed as $s_i = \frac{1}{M} \sum_{m=1}^{M} s_i^m$. The attention mask is then modified as follow:
\begin{align*}
    \text{Attn}(x^{d_i}) = s_i \times \text{Attn}(x^{d_i}) \times C,
\end{align*}
where $x^{d_i}$ is tokens that belongs to document $d_i$ and $C$
is normalization constant to ensure the total attention remains unchanged.

%% file: latex/sections/experiments.tex
\input{./latex/tables/exp_main}

\section{Experiment}
We evaluate the proposed method across a variety of question-answering datasets, aiming to validate the following claims about $\SYSNAME$: 
\begin{itemize}[noitemsep, topsep=0pt, leftmargin=*]
    \item \textit{Improving LLM performance with context information (Section\ref{sec:exp_main})}: $\SYSNAME$ strengthens LLM robustness when presented with context information of varying credibility.
    \item \textit{Robustness with increasing number of noisy documents (Section \ref{sec:exp_noise})}: $\SYSNAME$ performance improvement persists with increasing amount of noise.
    \item \textit{Ablations (Section\ref{sec:exp_ablation})}: $\SYSNAME$ benefits with using more number of embedding functions, and can effectively incorporate information from extra documents, up to a certain limit.
\end{itemize}

\subsection{Experimental Setup}
\paragraph{Models} We evaluate $\SYSNAME$ on three publicly available models with varying sizes: \textit{Mistral-7B-Insruct-v0.2} , \textit{Mistral-Nemo-Instruct-2407} \cite{jiang2023mistral}, and \textit{Gemma-7b-it}~\footnote{https://huggingface.co/google/gemma-7b-it}. 

\paragraph{Datasets and Retriever} We validate our approach on four short form generation datasets: HotpotQA (HPQA) \cite{yang2018hotpotqa}, PopQA \cite{mallen2023llm_memorization}, NaturalQuestions (NQ) \cite{47761}, TriviaQA (TQA) \cite{joshi-etal-2017-triviaqa}, and one long-form generation dataset, ASQA \cite{stelmakh2022asqa}. 

$\SYSNAME$ operates after the retrieval step and is not dependent on the choice of retrievers. Therefore, we use documents retrieved from previous open-source research. For HPQA, we use the dataset and retrieved documents from \cite{pan2024contexts}, which provides a varying number of documents per query (approximately 25 to 50). For PopQA, NQ, TQA, and ASQA, we use the dataset and retrieved documents from \citet{wei2024instructrag}, where each query is associated with five retrieved documents.

\paragraph{Metrics.} We measure performance using \textit{Accuracy}, determined by checking if the correct answer is included in the model's generated output, following the standard evaluation \cite{asai2023self, mallen-etal-2023-trust, schick2024toolformer}. We use the official metric to evaluate ASQA, namely correctness (\textbf{str-em}), and \textbf{F1} score derived from citation precision and recall.

\subsection{Main Results}
\label{sec:exp_main}
\paragraph{Setup.} We compare $\SYSNAME$,  including the black-box approach - \textbf{$\SYSNAME$-bbx} and the white-box appraoch - \textbf{$\SYSNAME$-wbx},  with Vanilla RAG and two RAG document denoising methods: InstructRAG (or InstrRAG) \cite{wei2024instructrag} and CLeHe \cite{qiu2024entropy}. These methods were chosen for comparison because (1) they do not rely on access to labels or human-annotated credibility, and (2) they require no additional tuning. InstructRAG prompts the RAG model to reason about each document’s relevance to the query, then uses this reasoning as in-context examples. CLeHe is a decoding-time method that re-weight document attention based on the uncertainty of each document-conditioned next-token distribution.

\paragraph{Results.} As shown in Table \ref{tab:exp_main}, 
$\SYSNAME$ achieves significant performance improvements over both the Vanilla baseline and existing RAG denoising methods. In addition to accuracy gains, $\SYSNAME$ offers better scalability with larger document sets, as it processes documents in parallel without requiring individual model queries. This offers a distinct advantage over methods like InstructRAG and CLeHe, which incur linear computational overhead due to the need for separate model prompts for document-specific reasoning and next-token distribution computations. Furthermore, $\SYSNAME$-wbx is fully compatible with prompting-based approaches such as InstructRAG, enabling seamless integration for enhanced performance.

\section{Dissecting \SYSNAME: Strengths and Failure Modes}
This section provides a detailed analysis of $\SYSNAME$, highlighting both its strengths and potential failure modes to better understand its overall efficacy.
\subsection{Robust with Increasing Number of Noisy Documents}
\input{./latex/figures/test_noise}
\label{sec:exp_noise}
\paragraph{Setup.} We artificially inject noise into the retrieved documents in PopQA by swapping the golden answer keyword in documents that contain them into another word. We ensure that the replacement word comes from the same category as the original word by swapping words only from the same category. We vary the amount of noise to 20, 40, 60, and 80 percent.

\paragraph{Results.} Figure \ref{fig:exp_noise_popqa} demonstrates that $\SYSNAME$-bbx exhibits superior noise robustness compared to InstructRAG, with a modest advantage on Mistral and Nemotron models and a substantial improvement on Gemma. While the combination of $\SYSNAME$-wbx and InstrRAG achieves optimal performance on larger models (Mistral and Nemotron), $\SYSNAME$-wbx alone performs better on Gemma. These results not only underscore $\SYSNAME$'s enhanced robustness in high-noise settings but also highlight a crucial insight: for smaller models, methods that rely on the model's inherent reasoning capabilities may introduce additional noise, potentially degrading performance. This suggests the importance of developing techniques that remain effective independent of model capacity.

\subsection{Ablation Study}
\label{sec:exp_ablation}
\input{./latex/figures/ablations}

\paragraph{Setup.} We investigate how two key parameters affect performance: (1) the number of embedding models used, and (2) the number of retrieved documents. We evaluate performance on the PopQA dataset while varying the number of embedding functions (embedders) from one to five and the number of retrieved documents from one to twenty.

\paragraph{Results.} Figure \ref{fig:test1} reveals that $\SYSNAME$-bbx achieves consistent accuracy improvements as we increase the number of embedder functions, though these gains plateau after seven embedders. In contrast, $\SYSNAME$-qbx shows no significant performance improvements from additional embedder models. Figure \ref{fig:test2} demonstrates that both variants benefit from increasing the number of retrieved documents, with performance improvements continuing up to approximately 12 documents. Beyond this point, we observe slight performance degradation: $\SYSNAME$-bbx experiences a 0.5\% accuracy drop after 12 documents, while $\SYSNAME$-wbx shows a 1.5\% decline after 18 documents. These results suggest that $\SYSNAME$ effectively leverages information from additional documents up to a certain threshold. However, as the document set grows larger, the proportion of documents containing correct answers diminishes, leading to gradual performance deterioration.

%% file: latex/tables/exp_main.tex
\begin{table*}[t!]
   \centering
   \begin{tabular}{llccccccccc}
     \toprule
    \multirow{2}{*}{Model} & \multirow{2}{*}{Method} &  \multirow{2}{*}{HPQA} &  \multirow{2}{*}{PopQA} &  \multirow{2}{*}{NQ} &  \multirow{2}{*}{TQA} & \multicolumn{2}{c}{ASQA} \\ 
      \cmidrule(lr){7-8}
    &&  &  &  &  & str-em & F1   \\
    \toprule
     \multirow{6}{*}{Mistral 7B}
     & Vanilla model & 43.00 & 60.30 & 59.66 & 74.06 & 39.43 & 31.95 \\
     & CLeHe \cite{qiu2024entropy} & 40.68 & 55.75 & 58.31 & 75.08 & 38.28 & \textbf{34.76} \\
     & InstructRAG \cite{wei2024instructrag} & 46.40 & 60.40 & 57.90 & 74.30 & 42.00 & 32.85 \\
     & \textbf{$\SYSNAME$--bbx}  & \underline{50.00} & \underline{63.47} & \underline{64.27} & \underline{75.85} & \underline{43.95} & 32.78 \\
     & \textbf{$\SYSNAME$--wbx}  & 48.36 & 61.09 & 61.36 & 74.60 & 41.33 & 30.12 \\
    \cmidrule{2-8} 
    & \textbf{$\SYSNAME$--wbx} + InstructRAG & \textbf{70.71} & \textbf{67.86} & \textbf{80.88} & \textbf{84.53} & \textbf{55.00} & \underline{34.72} \\ 
     
     \midrule

     \multirow{6}{*}{Mistral Nemo}
     & Vanilla model & 45.60 & 57.54 & 55.82 & 72.79 & 31.66 & 14.61 \\
     & CLeHe \cite{qiu2024entropy} & 36.84 & 42.46 & 54.92 & 72.15 & 30.36 & 22.60 \\
     & InstructRAG \cite{wei2024instructrag}& \underline{64.80} & \underline{71.30} & \underline{74.00} & \textbf{81.20} & 35.46 & \textbf{29.57} \\
     & \textbf{$\SYSNAME$--bbx}  & 56.38 & 61.90 & 65.87 & 78.32 & 36.32 & 28.78 \\
     & \textbf{$\SYSNAME$--wbx}  & 53.92 & 59.16 & 60.91 & 74.68 & \underline{36.86} & 26.82 \\
    \cmidrule{2-8} 
    & \textbf{$\SYSNAME$--wbx} + InstructRAG &  \textbf{72.47} & \textbf{79.78} & \textbf{84.72} & \underline{79.02} & \textbf{53.52} & \underline{24.38} \\

    \midrule

     \multirow{6}{*}{Gemma 7B}
     & Vanilla model & 26.60 & 48.82 & 42.66 & 58.75 & 26.21 & 23.40 \\
     & CLeHe \cite{qiu2024entropy} & 0.00 & 19.37 & 12.66 & 28.16 & 13.24 & 19.61 \\
     & InstructRAG \cite{wei2024instructrag} & 25.60 & 36.45 & \underline{59.06} & 46.87 & \underline{43.86} & \textbf{31.93} \\
     & \textbf{$\SYSNAME$--bbx}  & 22.60 & \textbf{70.32} & \textbf{73.81} & \textbf{79.20} & \textbf{58.62} & 15.73 \\
     & \textbf{$\SYSNAME$--wbx}  & \underline{36.82} & \underline{55.94} & 45.60 & \underline{57.92} & 29.86 & \underline{26.63}  \\
    \cmidrule{2-8} 
    & \textbf{$\SYSNAME$--wbx} + InstructRAG & \textbf{44.24} & 31.39 & 49.03 & 40.41 & 29.54 & 20.82 \\ 

    \bottomrule
   \end{tabular}

   \caption{$\SYSNAME$ credibility score improves vanilla model and outperforms strong baselines, despite having no access to manual credibility annotations and model-derived signals. Best results in \textbf{bold} and second best \underline{underlined}.}

   \label{tab:exp_main}
 \end{table*}

%% file: latex/figures/test_noise.tex
\begin{figure*}[ht!]

\centering
\includegraphics[width=.325\textwidth]{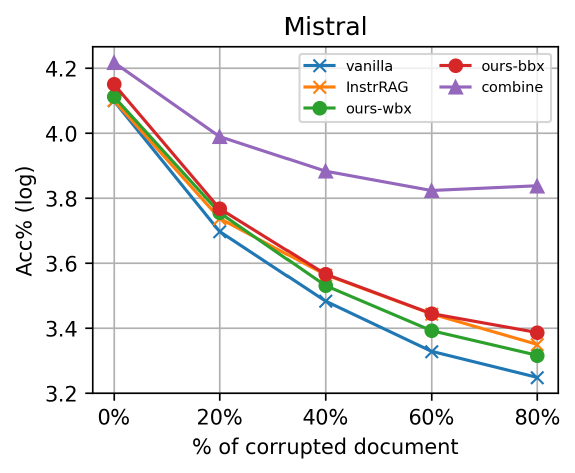}
\includegraphics[width=.325\textwidth]{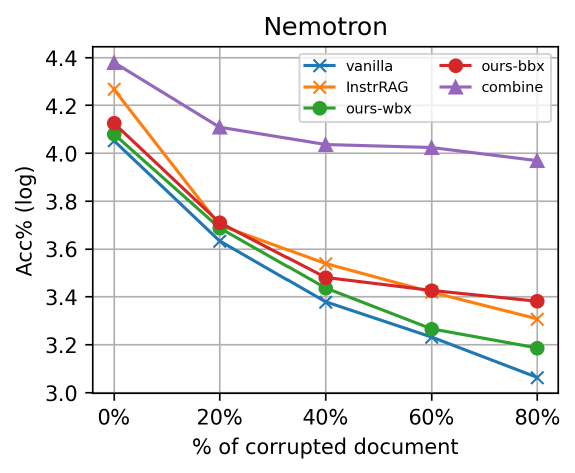}
\includegraphics[width=.325\textwidth]{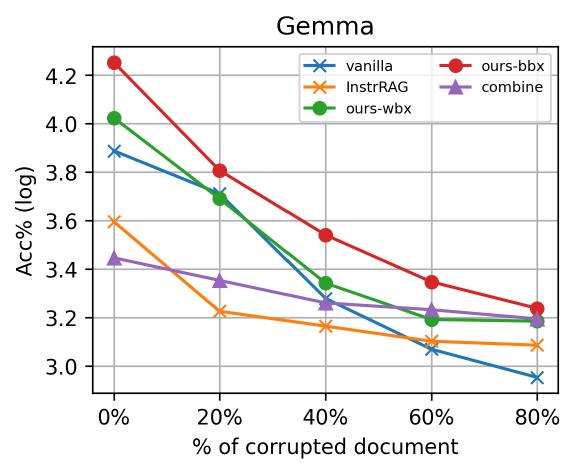}

\caption{$\SYSNAME$ with increasing noise. X-axis: \% of corrupted documents, Y-axis: Accuracy \% (log scale). }
\label{fig:exp_noise_popqa}

\end{figure*}

%% file: latex/figures/ablations.tex
\begin{figure*}[ht!]
\centering
\begin{minipage}{.43\textwidth}
  \centering
  \includegraphics[width=\linewidth]{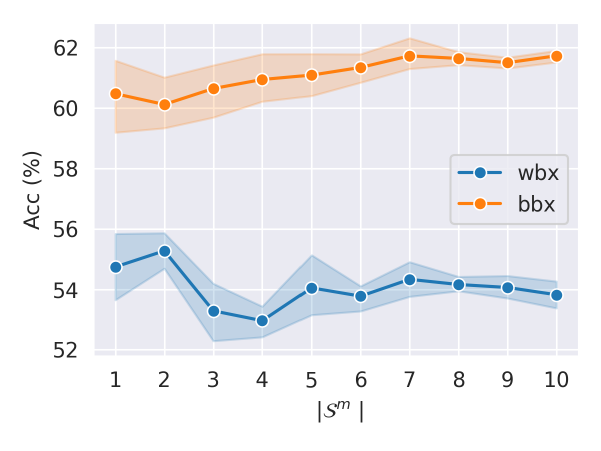}
  \captionof{figure}{Impact of the number of embedder functions. $\SYSNAME$-bbx benefits from using more embedders.}
  \label{fig:test1}
\end{minipage}%
\hfill
\begin{minipage}{.43\textwidth}
  \centering
  \includegraphics[width=\linewidth]{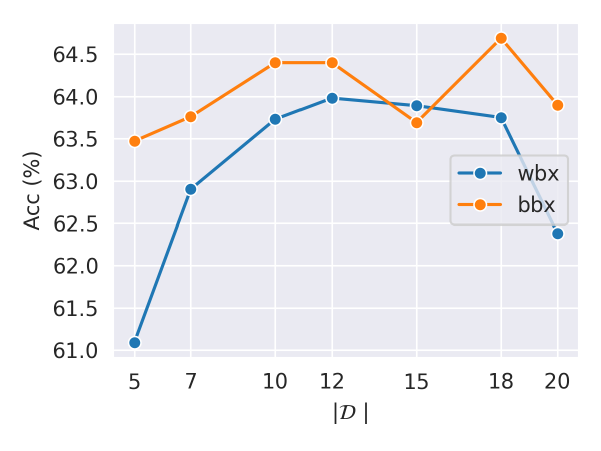}
  \captionof{figure}{Impact of the number of retrieved documents. Both $\SYSNAME$-wbx and bbx efficacy persist with increasing number of documents.}
  \label{fig:test2}
\end{minipage}
 \vspace{-1em}
\end{figure*}

%% file: latex/sections/analysis.tex
\input{./latex/figures/score_dist}
\subsection{How does the $\SYSNAME$ credibility score distribution look like?}
Figure \ref{fig:analysis_score_dist} shows the cumulative distribution of CrEST scores for golden and distractor documents. The clear separation between the cumulative distributions indicates that CrEST effectively assigns different score ranges to golden and distractor documents. Additionally, the steeper slope of the distractor curve around score 0.0 indicates a concentration of distractor documents receiving neutral scores, while golden documents show a more gradual distribution skewed toward positive scores. This aligns with our expectation that documents containing correct answers should exhibit stronger semantic coherence with the retrieved document set.

\input{./latex/figures/docs_distance}
\subsection{How Does Noise Level Affect Pairwise Document Similarity?}
Figure \ref{fig:analysis_dist_noise} illustrates the relationship between noise levels and average pairwise distances among golden documents (blue), corrupted documents (orange), and the complete document set (dashed green). At low noise levels, golden documents exhibit substantially smaller inter-document distances, enabling more accurate credibility estimation and improved system performance (Figure \ref{fig:exp_noise_popqa}). However, as noise levels increase, the distance distributions between corrupted and golden documents gradually converge, reducing their discriminability and undermining the reliability of credibility estimates. This convergence pattern provides mechanistic insight into $\SYSNAME$'s performance deterioration under high-noise conditions.

\input{./latex/figures/att_dist}
\subsection{How Does Attention Distribution Change Before and After $\SYSNAME$-wbx?}
Figure \ref{fig:attn_analysis} illustrates the distribution of normalized attention scores across document tokens, comparing the vanilla model (left, orange) with $\SYSNAME$-wbx (right, blue). Without $\SYSNAME$-wbx, most tokens receive uniform attention scores of 1, preventing the model from distinguishing between important and irrelevant tokens. In contrast, $\SYSNAME$-wbx redistributes attention weights to emphasize tokens based on their origin document estimated credibility. This redistribution allows the model to emphasize informative tokens more while reducing the influence of potentially non-credible content.

\input{./latex/figures/analysis_bbx}
\subsection{How Distinguishable Are the Outputs from $\SYSNAME$-bbx in the Latent Space?}
We analyze the effectiveness of $\SYSNAME$-bbx's LLM output aggregation mechanism (described in Section \ref{sec:black_white_box}) by examining how well it identifies correct answers in the latent space. Figure \ref{fig:bbx-analysis} presents the frequency distribution of correct answer ranks based on pairwise distances. The results reveal that in approximately 85\% of cases, the correct answer achieves the lowest pairwise distance (Rank 1), indicating that $\SYSNAME$-bbx successfully identifies it as the most reliable response. When the method doesn't identify the correct answer as the top choice, it typically ranks it second (Rank 2), with a sharp decline in frequency for lower ranks. This strong bias toward top rankings demonstrates that even when $\SYSNAME$-bbx doesn't perfectly identify the correct answer, it still maintains it within a close distance in the latent space.

%% file: latex/figures/score_dist.tex
\begin{figure}[ht!]
    \centering
    \includegraphics[width=.7\linewidth]{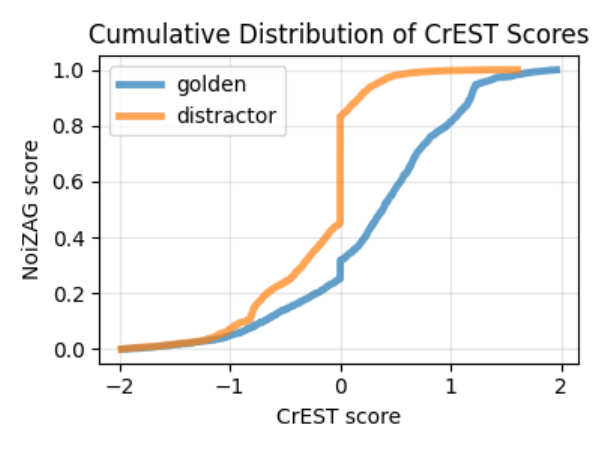}
    \caption{$\SYSNAME$ score distribution on NaturalQuestions (NQ) for both documents containing the golden answer (blue) and distractor documents (orange).}
    \label{fig:analysis_score_dist}
    \vspace{-1em}
\end{figure}

%% file: latex/figures/docs_distance.tex
\begin{figure}[ht!]
    \centering
    \includegraphics[width=.73\linewidth]{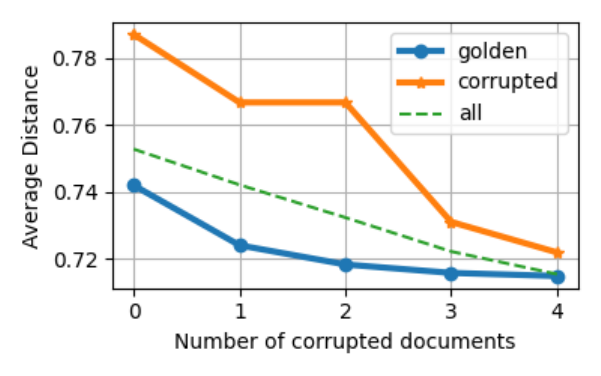}
    \caption{Avg. distance between golden (orange) and corrupted (blue) documents with increasing number of corrupted documents.}
    \label{fig:analysis_dist_noise}
     \vspace{-1em}
\end{figure}

%% file: latex/figures/att_dist.tex
\begin{figure}[ht!]
    \centering
    \includegraphics[width=0.9\linewidth]{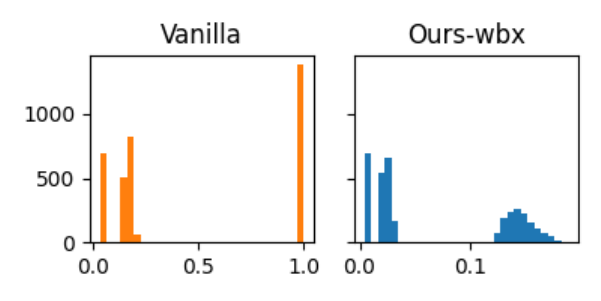}
    \caption{Attention score distribution before $\SYSNAME$-wbx (orange) and after (blue). $\SYSNAME$ re-distributes the attention score of each documents token based on the document's credibility.}
    \label{fig:attn_analysis}
     \vspace{-0.5em}
\end{figure}

%% file: latex/figures/analysis_bbx.tex
\begin{figure}[ht!]
    \centering
    \includegraphics[width=.65\linewidth]{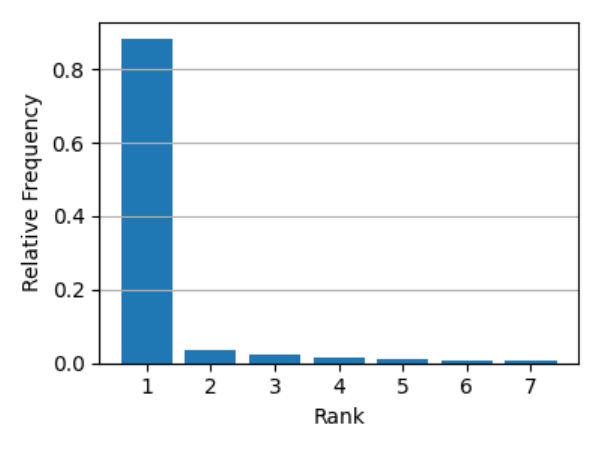}
    \caption{Frequency distribution of the correct answer's rank in $\SYSNAME$-bbx. We measure how often the correct answer has the lowest pairwise distance (Rank 1), the second lowest (Rank 2), and so on.}
    \label{fig:bbx-analysis}
     \vspace{-0.5em}
\end{figure}

%% file: latex/sections/related_work.tex
\section{Related Work}
\paragraph{Context-Aware Generation.}
Context-aware text generation is a well-studied area, with in-context learning and retrieval-augmented generation being prominent techniques. These methods are widely employed to enhance LLM performance in knowledge-intensive tasks \cite{lewis2020retrieval}. By incorporating supplementary information, such as retrieved passages from external knowledge sources, these approaches improve performance by using the additional context as in-context examples to enable grounded inference. Recent efforts have focused on further optimizing this approach with various fine-tuning techniques. For example, \cite{asai2023self} trains LLMs to selectively use retrieval, \cite{luo2023search, wei2024instructrag} applies instruction-tuning with documents as in-context examples, and \cite{gautier2022few} jointly fine-tunes both the LLM and the document retriever. While these methods enhance performance, they also increase computational costs and mainly focus on improving the model's ability to utilize the provided documents. In this work, we propose a computationally efficient, \textit{weakly-supervised approach that assesses the credibility of retrieved documents}, allowing the LLM to avoid distractions from noisy or non-credible sources when generating answers.

\paragraph{Incorporating Noisy Context Information.} Recent research \cite{pmlr-v202-shi23a} has shown that LLMs can be easily distracted by irrelevant context, prompting a series of studies on the impact of irrelevant and misleading documents. For instance, \cite{zhang2024raft} proposed fine-tuning LLMs to understand the relationship between the query, retrieved documents, and the correct answer—encouraging the model to memorize answers when no relevant document is available. \cite{pan2024contexts} fine-tuned models using human-annotated document credibility to help distinguish credible from non-credible documents. Similarly, \cite{deng2024cram} employed off-the-shelf re-ranker models to adjust attention masks during inference, ensuring the model focuses more on tokens from relevant documents. Additionally, \cite{yan2024corrective} derived heuristic relevance scores to fine-tune LLMs. However, these methods rely on human annotations, either as labels or credibility information, limiting their scalability and practicality for large datasets. To reduce this requirement, \cite{wei2024instructrag} introduced an in-context learning approach, where the model first assesses document relevance, and \cite{qiu2024entropy} re-weighted documents during decoding using probability distributions.  Our approach aligns with these efforts, especially the last two, but goes further by eliminating the need to query the LLM for each document, enabling the use of larger document sets per query. \cite{cuconasu2024power} showed a surprising result where randomly selected, irrelevant documents can actually improve performance when placed carefully within a prompt. Our study is orthogonal to their work and can potentially incorporate their finding for further improvements. 

\paragraph{Weak Supervision}
$\SYSNAME$ uses weak supervision to estimate document credibility without relying on explicit labels or human annotations. Traditionally, weak supervision frameworks automatically generate binary labels by aggregating outputs from multiple noisy labeling functions, estimating each function's accuracy through inter-function agreement rates \cite{ratner2016data, ratner2017snorkel, fu2020fast}. This approach has been extended to handle structured outputs \cite{shin2021universalizing}, enabling the estimation of continuous values. The effectiveness of weak supervision has been demonstrated across diverse applications, including image segmentation, chemical reaction extraction, and information retrieval \cite{hooper2020cut, guha2024smoothie, liu2017heterogeneous}. Building on this foundation, we adapt the weak supervision paradigm for a novel purpose: assessing the credibility of retrieved documents in the context of language model inference.

%% file: latex/sections/conclusion.tex
\section{Conclusion}
In this paper, we introduced $\SYSNAME$, a novel approach that enhances the use of contextual information in LLMs by estimating document credibility via weak supervision—eliminating the need for manual labels, extensive fine-tuning, or reliance on model reasoning. $\SYSNAME$ filters noisy content by prioritizing widely supported information, enabling LLMs to focus on credible context. Future work includes training retrievers with $\SYSNAME$’s loss and exploring stronger white-box integration methods. Overall, $\SYSNAME$ offers a practical and effective way to improve robustness in context-aware generation.

%% file: latex/appendix/glossary.tex
\section{Glossary} 
We list definition of mathematical notation used in this paper in Table \ref{table:glossary}
\label{appendix:glossary}
\label{sec:gloss}

\begin{table}[htp!]
\centering
\small
\begin{tabular}{l l}
\toprule
Symbol & Definition \\
\midrule
$x$ & User query \\
$d_i$ & A single retrieved document \\
$\mathcal{D}$ & A set of retrieved documents for query $x$ \\
$\mathcal{M}$ & Generator LLM \\
$s_i$ & Reliability score for document $d_i$ \\
$\mathcal{S}$ & Set of reliability scores of the retrieved documents \\
$\lambda^i$ & Embedding of document $d_i$ \\
$\lambda^*$ & Unobserved true document embedding \\
$\hat{y}$ & LLM prediction \\
\bottomrule
\end{tabular}
\caption{
	Glossary of variables and symbols used in this paper.
}
\label{table:glossary}
\end{table}

%% file: latex/appendix/eq3_derivation.tex
\section{Triplet Model Derivation}
\label{sec:triplet_derivation}
Pick 3 document embeddings $\lambda_i, \lambda_j, \lambda_k$, and plug it in equation \ref{eq:exppectation} we have:
\begin{align}
   & \mathbbm{E}[\| \lambda_i - \lambda_j \|^2] = \mathbbm{E}[\| \lambda_i - \lambda^* \|^2] + \mathbbm{E}[\| \lambda_j - \lambda^* \|^2] \\
   & \mathbbm{E}[\| \lambda_i - \lambda_k \|^2] = \mathbbm{E}[\| \lambda_i - \lambda^* \|^2] + \mathbbm{E}[\| \lambda_k - \lambda^* \|^2] \\
   & \mathbbm{E}[\| \lambda_j - \lambda_k \|^2] = \mathbbm{E}[\| \lambda_j - \lambda^* \|^2] + \mathbbm{E}[\| \lambda_k - \lambda^* \|^2]
\end{align}
Now define:
\begin{align*}
  &  \delta_{ij} =  \mathbbm{E}[\| \lambda_i - \lambda_j \|^2], \quad \delta_{ik} =  \mathbbm{E}[\| \lambda_i - \lambda_k \|^2], \quad \delta_{jk} =  \mathbbm{E}[\| \lambda_j - \lambda_k \|^2]
\end{align*}
Adding the first two equations (5 and 6), we get:
\begin{align*}
    & \delta_{ij} + \delta_{ik} = 2 \mathbbm{E}[\| \lambda_i - \lambda^* \|^2]  + \mathbbm{E}[\| \lambda_j - \lambda^* \|^2]  +  \mathbbm{E}[\| \lambda_k - \lambda^* \|^2] 
\end{align*}
Subtract this with the third equation (7), and divide by 2; we get:
\begin{align*}
    & \frac{1}{2} \delta_{ij} + \delta_{ik} - \delta_{jk} = \mathbbm{E}[\| \lambda_i - \lambda^* \|^2]
\end{align*}
This is Equation \ref{eq:triplet}.

%% file: latex/appendix/dataset_details.tex
\subsection{Dataset Details} 
\label{appendix:dataset_details}

\begin{table}[h]
\centering
\small
\begin{tabular}{l l l}
\toprule
Dataset & Number of Test samples & Retriever \\
\midrule
HotpotQA & 500 & SPLADE \\
PopQA & 1,399 & Contriever \\
NaturalQuestions &  3,610 & DPR  \\
TriviaQA &  11,313 & Contriever \\
ASQA & 948 & GTR \\
\bottomrule
\end{tabular}
\caption{Dataset and retriever details}
\label{table:dataset}
\end{table}
We include dataset details in table \ref{table:dataset}, including the number of test samples and retriever used in the setup we follow from \cite{wei2024instructrag} and \cite{pan2024contexts}.

%% file: latex/appendix/embedder_details.tex
\section{Embedder Details}
\label{appendix:embedder_models}
We use the following embedding models to compute document similarity (distances) for reliability scores:

\begin{enumerate}
    \item nvidia/NV-Embed-v1 \cite{lee2024nvembed}
    \item SIMCSE \cite{gao2021simcse}
    \item infgrad/stella\_en\_1.5B\_v5
    \item ms-marco reranker models \cite{reimers-2019-sentence-bert}
    
\end{enumerate}